\documentclass{article}

    \PassOptionsToPackage{numbers}{natbib}

\usepackage[preprint]{neurips_2025}

\usepackage[utf8]{inputenc} 
\usepackage[T1]{fontenc}    
\usepackage{hyperref}       
\usepackage{url}            
\usepackage{booktabs}       
\usepackage{amsmath}
\usepackage{amssymb}
\usepackage{amsfonts}       
\usepackage{nicefrac}       
\usepackage{microtype}      
\usepackage[svgnames]{xcolor}         
\usepackage{cleveref}
\usepackage{graphicx}
\usepackage{algorithm}
\usepackage{algorithmic}
\usepackage{orcidlink}
\usepackage{dsfont}

\definecolor{blue4}{HTML}{0000FF}

\usepackage{fontawesome5}
\definecolor{blue4}{HTML}{0000FF}
\definecolor{riceblue}{HTML}{00205B}

\hypersetup{
    colorlinks=true,       
    linkcolor=blue4,          
    citecolor=blue4,        
    urlcolor=blue4           
}

\title{FlowVAT: Normalizing Flow Variational Inference with Affine-Invariant Tempering}

\author{%
Juehang Qin \orcidlink{0000-0001-8228-8949}\\ 
Rice University \\
\texttt{qinjuehang@rice.edu} 
\And Shixiao Liang \orcidlink{0000-0003-0116-654X}\\
Rice University \\
\texttt{liangsx@rice.edu} 
\And Christopher Tunnell \orcidlink{0000-0001-8158-7795}\\ 
Rice University \\
\texttt{tunnell@rice.edu}
}

\begin{document}

\maketitle

\begin{abstract}


Multi-modal and high-dimensional posteriors present significant challenges for variational inference, causing mode-seeking behavior and collapse despite the theoretical expressiveness of normalizing flows. Traditional annealing methods require temperature schedules and hyperparameter tuning, falling short of the goal of truly black-box variational inference. We introduce FlowVAT, a conditional tempering approach for normalizing flow variational inference that addresses these limitations. Our method tempers both the base and target distributions simultaneously, maintaining affine-invariance under tempering. By conditioning the normalizing flow on temperature, we leverage overparameterized neural networks' generalization capabilities to train a single flow representing the posterior across a range of temperatures. This preserves modes identified at higher temperatures when sampling from the variational posterior at $T = 1$, mitigating standard variational methods' mode-seeking behavior. In experiments with 2, 10, and 20 dimensional multi-modal distributions, FlowVAT outperforms traditional and adaptive annealing methods, finding more modes and achieving better ELBO values, particularly in higher dimensions where existing approaches fail. Our method requires minimal hyperparameter tuning and does not require an annealing schedule, advancing toward fully-automatic black-box variational inference for complicated posteriors.

\end{abstract}

\section{Introduction}\label{sec:intro}

Statistical inference forms the cornerstone of empirical research, allowing us to draw meaningful conclusions about unobserved variables from observed data~\cite{gelman1995bayesian}. This process is frequently framed in terms of posterior inference, where we have an observed dataset, $\boldsymbol{X}$, and aim to obtain a posterior distribution, $p(\boldsymbol{\theta}|\boldsymbol{X})$, over parameters $\boldsymbol{\theta}$, by combining a likelihood function $p(\boldsymbol{X}|\boldsymbol{\theta})$ with a prior distribution $p(\boldsymbol{\theta})$. Since exact inference is often intractable, researchers have developed various approximate methods. Variational Inference (VI) has emerged as a particularly powerful approach~\cite{gelman1995bayesian, Ranganath2013BlackBV, rezende2015variational}, wherein the posterior is approximated by a parameterized family of distributions, allowing us to find an approximate distribution $q_\phi(\boldsymbol{\theta})$ that closely resembles the true posterior $p(\boldsymbol{\theta}|\boldsymbol{X})$.

A significant advancement in this field has been the development of black-box variational inference~\cite{Ranganath2013BlackBV, rezende2015variational, Giordano2023BlackBV}, which provides a general-purpose framework applicable to diverse inference problems without requiring tailored variational families or methodologies for each specific scenario. When combined with highly expressive distribution families such as normalizing flows, these methods can theoretically represent even complex multimodal posteriors~\cite{rezende2015variational, agrawal2020advances}. However, despite the theoretical expressiveness of these models, variational methods in practice often exhibit mode-seeking behavior~\cite{soletskyi2024theoreticalperspectivemodecollapse, dhaka2021challenges}, presenting a significant obstacle to the development of fully-automated variational inference systems capable of handling high-dimensional posteriors with multiple modes. Such high-dimensional and multimodal posteriors are often encountered in science, including in neutrino oscillation analyses~\cite{esteban2018updated, coloma2023global}, searches for physics beyond the Standard Model~\cite{Fowlie:2011mb}, and gravitational wave inference~\cite{Dax:2024mcn}.

To address these challenges, researchers have explored tempering approaches that can improve inference for complicated posteriors. Various tempering strategies and schedules have been investigated for both variational methods and Markov Chain Monte Carlo (MCMC) sampling ~\cite{Mandt2014VariationalT, SwendsenPhysRevLett.57.2607, Surjanovic2022ParallelTW}. The fundamental idea behind tempering is to modify either the likelihood or the posterior by raising it to the power of $1/T$, creating a shallower distribution that is easier to fit. The temperature parameter $T$ can then be decreased until we recover the original target distribution at $T=1$.

In this work, we introduce a novel approach that eliminates the need for annealing schedules and associated hyperparameters, bringing us one step closer to fully-automatic black-box variational inference capable of handling complicated multimodal posteriors. Our key insight is that by simultaneously tempering both the base distribution and the target posterior, the transformation represented by the normalizing flow can remain unchanged even as temperature changes. This means that the transformation represented by the normalizing flow does not need to vary as much across temperatures, thus promoting learning transformations that are generalizable across temperatures. Furthermore, by conditioning the normalizing flow on the temperature parameter, we leverage the generalization capabilities of overparameterized neural networks ~\cite{wilson2025deeplearningmysteriousdifferent, schaeffer2023doubledescentdemystifiedidentifying} to train a single flow that can accurately represent the posterior distribution across a wide spectrum of temperatures. Crucially, this approach preserves modes identified at higher temperatures even at $T=1$, thereby addressing one of the fundamental limitations of traditional variational methods. We call our method \textbf{FlowVAT}: normalizing \textbf{Flow} \textbf{V}ariational inference with \textbf{A}ffine-invariant \textbf{T}empering, reflecting its integration of normalizing flows with our novel affine-invariant tempering approach to enhance variational inference. Code used to generate all data and figures in this paper will be made available on github.


\section{Affine-Invariant Tempering for Normalizing Flow Variational Inference}\label{sec:flowvat}

There are two key motivations for our approach. First, we aim to leverage the demonstrated capacity of overparameterized neural networks to generalize effectively~\cite{wilson2025deeplearningmysteriousdifferent, schaeffer2023doubledescentdemystifiedidentifying}. Second, we seek to introduce a beneficial inductive bias that enhances inference quality. This is accomplished by applying temperature scaling to both the base distribution and the posterior distribution of a normalizing flow.

A normalizing flow, represented by the transformation $\boldsymbol{f}_\phi$, defines a probability distribution through:
\begin{equation}\label{eq:norm_flow}
    q_\phi(\boldsymbol{\theta}) = q(\boldsymbol{z})\left|\frac{\partial \boldsymbol{f}_\phi}{\partial \boldsymbol{z}}\right|^{-1},
\end{equation}
where $\phi$ denotes the weights and parameters of the normalizing flow, $\boldsymbol{z}$ represents the space of the base distribution, and $\boldsymbol{\theta}$ corresponds to the parameter space for inference. In variational inference, we optimize the normalizing flow using loss function that is minimized when the approximating distribution $q_\phi(\boldsymbol{\theta})$ converges to the true posterior $p(\boldsymbol{\theta}|\boldsymbol{X})$, such as the negative evidence lower bound (ELBO)~\cite{Hoffman2019arXiv190303704H, Ranganath2013BlackBV}. Maximizing the ELBO is equivalent to minimizing the Kullback-Leibler (KL) divergence~\cite{kingma2013auto}. The ELBO can be expressed as:
\begin{equation}\label{eq:elbo}
    \mathcal{L}(\phi) = \mathbb{E}_{q(\boldsymbol{z})}\left[\log\left(p(\boldsymbol{X}, \boldsymbol{f}_\phi (\boldsymbol{z}))\right) - \log\left(q(\boldsymbol{z})\left|\frac{\partial \boldsymbol{f}_\phi}{\partial \boldsymbol{z}}\right|^{-1}\right)\right]\,.
\end{equation}

\subsection{Theoretical Framework and Heuristic Arguments}

To illustrate why tempering both the base and transformed distributions improves training, we examine a simple case where the base distribution is a standard Gaussian, $\mathcal{N}(\mu = 0, \sigma = \mathds{1})$, and the posterior is a multivariate Gaussian. In such instances, the optimal transformation for a normalizing flow would be an affine transformation, as this can transform a standard Gaussian distribution into an an arbitrary multivariate Gaussian:
\begin{equation}\label{eq:affine}
    \boldsymbol{\theta} = \boldsymbol{f}(z) = A\boldsymbol{z} + \boldsymbol{b}.
\end{equation}
The Jacobian of this transformation is the matrix $A$, resulting in the log determinant of the Jacobian matrix, $\log\left|{\partial \boldsymbol{f}}/{\partial \boldsymbol{z}}\right|$ being a constant. From~\cref{eq:elbo}, we observe that a constant value of $\log\left|{\partial \boldsymbol{f}}/{\partial \boldsymbol{z}}\right|$ does not affect the gradient of the ELBO.

When training with a tempered posterior, we raise the original posterior to a power of $1/T$~\cite{Mandt2014VariationalT}, yielding:
\begin{equation}
    p^{1/T}(\boldsymbol{\theta}|\boldsymbol{X}) = \frac{p^{1/T}(\boldsymbol{X}, \boldsymbol{\theta})}{\int p^{1/T}(\boldsymbol{X}, \boldsymbol{\theta}) d\boldsymbol{\theta}}.
\end{equation}
This modifies our ELBO target to:
\begin{equation}
    \mathcal{L}_T(\phi) = \mathbb{E}_{q_T(\boldsymbol{z})}\left[\log\left(p^{1/T}(\boldsymbol{\theta}|\boldsymbol{X})\right) + \log\left(\int p^{1/T}(\boldsymbol{X}, \boldsymbol{\theta}) d\boldsymbol{\theta}\right)- \log\left(q_{T}(\boldsymbol{z})\left|\frac{\partial \boldsymbol{f}_{\phi,T}}{\partial \boldsymbol{z}}\right|^{-1}\right)\right].
\end{equation}

Since constant terms such as the marginalization integral do not affect the gradient, we can simplify to:
\begin{equation}
    \mathcal{L}_T'(\phi) = \mathbb{E}_{q_T(\boldsymbol{z})}\left[\log\left(p^{1/T}(\boldsymbol{\theta}|\boldsymbol{X})\right) - \log\left(q_{T}(\boldsymbol{z})\left|\frac{\partial \boldsymbol{f}_{\phi,T}}{\partial \boldsymbol{z}}\right|^{-1}\right)\right].
\end{equation}

By additionally tempering the base distribution as $q_T(\boldsymbol{z}) = q^{1/T}(\boldsymbol{z})$, we obtain:
\begin{equation}
    \mathcal{L}_T'(\phi) =\mathbb{E}_{q^{1/T}(\boldsymbol{z})}\left[\log\left(p^{1/T}(\boldsymbol{f}_{\phi,T}(z)|\boldsymbol{X})\right) - \log\left(q^{1/T}(\boldsymbol{z})\right) - \frac{1}{T}\log\left|\frac{\partial \boldsymbol{f}_{\phi,T}}{\partial \boldsymbol{z}}\right|^{-1}\right].
\end{equation}

For affine transformations as defined in~\cref{eq:affine}, the ELBO remains invariant except for the sampling distribution and constant or multiplicative factors. Importantly, if there exists a transformation for which the transformed posterior is exactly matched by the approximate distribution $q$ (indicating a Gaussian posterior when using a standard Gaussian base distribution), then the transformation $\boldsymbol{f}_{\phi}(z)$ that achieves $\log\left(p^{1/T}(\boldsymbol{f}_{\phi}(z)|\boldsymbol{X})\right) = \log\left(q^{1/T}(\boldsymbol{z})\right)$ does so regardless of temperature.

Furthermore, we can represent a non-affine transformation as the composition of an affine transformation and an additional non-affine component. While our tempering approach still requires the non-affine component to adapt with temperature, the affine transformation can remain invariant. Heuristically, this affine-invariant tempering promotes consistent matching of the first two moments (mean and variance) across temperatures, particularly when using overparameterized neural networks that exhibit strong generalization properties.

Matching the first two moments across temperatures provides an additional benefit. In practice, the ELBO is estimated using Monte Carlo (MC) samples; as such, regions of parameter space with low sampling probability may become effectively inaccessible. This can result in local minima in the loss function when separated modes are not adequately sampled. When means and variances are matched across temperatures, our dual tempering approach promotes a sampling distribution that maintains approximately the same mean position across temperatures in the parameter space, but with a wider distribution at higher temperatures. This significantly reduces the probability of becoming trapped in local minima caused by inadequate coverage of the parameter space by the sampling distribution.


\subsection{Evidence calculation using importance sampling}

Using temperature as a conditional variable also enhances estimation of Bayesian evidence via importance sampling. In variational inference, we typically do not have access to the normalized posterior $p(\boldsymbol{\theta}|\boldsymbol{X})$, but rather to an unnormalized function proportional to the posterior density, given by the product of the likelihood function and the prior:

\begin{equation}
    p'(\boldsymbol{\theta}) = p(\boldsymbol{X}|\boldsymbol{\theta})p(\boldsymbol{\theta}).
\end{equation}

The Bayesian evidence (marginal likelihood) is then given by:

\begin{equation}\label{eq:evidence_integral}
    \mathcal{Z} = \int p'(\boldsymbol{\theta})d\boldsymbol{\theta}.
\end{equation}

The Bayesian evidence is particularly useful for model comparison using Bayes factors~\cite{gelman1995bayesian}. However, the evidence integral can be high-dimensional and thus intractable. There are various approaches to estimating the evidence integral, such as Nested Sampling~\cite{Ashton:2022grj, skilling2004, skilling2006}. Another approach is to formulate this integral as an importance sampling problem~\cite{Llorente2020MarginalLC}, such that

\begin{equation}\begin{split}\label{eq:IS_evidence}
    \mathcal{Z} = \int q_\phi(\boldsymbol{\theta})\frac{p'(\boldsymbol{\theta})}{q_\phi(\boldsymbol{\theta})} d\boldsymbol{\theta}
    = \mathbb{E}_{q_\phi(\boldsymbol{\theta})}\left[\frac{p'(\boldsymbol{\theta})}{q_\phi(\boldsymbol{\theta})}\right],
\end{split}\end{equation}

where $q_\phi(\boldsymbol{\theta})$ is a normalized sampling distribution, in our case represented by the variational distribution. In practice, this approach is challenging, as it requires access to a proposal distribution that is normalized and covers the posterior sufficiently well while also remaining sufficiently concentrated. This is because if there are any regions of parameter space which are poorly covered by our approximate distribution, $q_\phi(\boldsymbol{\theta})$, then the importance ratio can have a heavy right tail; this is the motivation behind strategies such as Pareto Smoothed Importance Sampling~\cite{vehtari2024pareto}. On the other hand, using a very wide sampling distribution, such as the prior, would avoid a heavy right tail but would be extremely inefficient if the posterior is much more concentrated than the prior~\cite{Llorente2020MarginalLC}.

As FlowVAT results in a trained normalizing flow model with a conditional temperature parameter, we can avoid both of these issues by using the FlowVAT approximate posterior as a proposal distribution with. We can stabilize the importance ratios in~\cref{eq:IS_evidence} using $T>1$, thus decreasing the ratio $p'(\boldsymbol{\theta})/q_\phi(\boldsymbol{\theta})$ in the tails of the distribution. We demonstrate this approach in~\cref{sec:experiments}. For particularly challenging posteriors, it is possible to combine this approach of a tempered sampling distribution with Pareto Smoothed Importance Sampling to improve the reliability of evidence estimates, and the Pareto shape diagnostic can be used to determine whether estimates are reliable.

\section{Related Work}

One key approach to mitigate mode-seeking behaviour in variational inference has been the use of loss functions which promote mode-covering behavior, such as R\'enyi Divergence~\cite{Li2016RnyiDV}, which generalizes the divergence used in Importance Weighted Autoencoders~\cite{Burda2015ImportanceWA}, tail-adaptive $f$-divergence~\cite{Wang2018VariationalIW}, and the scale-invariant Alpha-Beta divergence~\cite{regli2018alphabetadivergencevariationalinference}. However, in high-dimensional problems, due to the instability of mode-covering divergences, this approach often does not improve performance~\cite{dhaka2021challenges}.

Another approach is to use tempered distributions. This approach with a linear annealing schedule was originally recommended when normalizing flows were introduced for variational inference~\cite{rezende2015variational}. AdaAnn~\cite{cobian2023adaann}, which we compare our work against, has been proposed more recently as an adaptive scheduler that adjusts the annealing schedule based on the change of the KL divergence. Relatedly, the approach of using temperature as a conditional variable has been explored in variational inference using a finite set of temperatures~\cite{Mandt2014VariationalT}. More recently, Annealing Flow~\cite{wu2024annealingflowgenerativemodel} proposed a continuous normalizing flow model guided by annealed intermediate distributions. While these methods share our motivation of addressing multi-modal posteriors through tempering, our approach differs by simultaneously tempering both base and target distributions in an affine-invariant manner, allowing for more stable training across temperature scales, and by eschewing the use of an annealing schedule.

Our approach is also related to Temperature-Steerable Flows \cite{dibak2022temperature}, which uses transformations with constant Jacobians to create normalizing flow models for sampling physical systems across thermodynamic states. However, our work focuses more generally on variational inference applications, and we rely on the flow to learn to generalize across temperatures while avoiding temperature dependence in the affine component, and expose temperature as a conditional variable to the neural network.

Normalizing flow models have also been used to accelerate other inference paradigms. This includes using flows as transport maps to accelerate MCMC~\cite{Hoffman2019arXiv190303704H} and nested sampling~\cite{amaral2025fast} by reparameterizing the problem to simplify the distribution being sampled, or as proposal distributions for importance nested sampling~\cite{lange_nautilus}. Our work focuses on variational inference, though these efforts could be viewed as complementary as improved variational posteriors that could potentially be used in conjunction with these methods. The use of normalizing flow models in conjunction with other approaches to posterior sampling also extends to estimations of Bayesian evidence, either using importance sampling of the harmonic mean estimator~\cite{Polanska:2024arc}, or using additional terms in the loss function to train a specialized normalizing flow and only using samples within the bulk of the posterior~\cite{Srinivasan:2024uax}.

Finally, our approach shares conceptual similarities with parallel tempering or replica-exchange MCMC sampling \cite{SwendsenPhysRevLett.57.2607, Marinari1992SimulatedTA, Earl2005ParallelTT, Surjanovic2022ParallelTW}, where multiple MCMC chains are instantiated at different temperatures, with communication between chains ensuring that those at low temperature can traverse regions with low probability density. While these methods operate in the MCMC framework, our work translates similar temperature-based intuitions to the variational setting, leveraging the generalization capabilities of neural networks to simultaneously represent the posterior across the entire temperature spectrum, and using the simultaneous tempering of the base and target distributions to promote positive inductive biases that allow variational posteriors at different temperatures to interact.

\section{Experiments}\label{sec:experiments}

In our experiments, we compare FlowVAT, where we temper both the posterior and the base distribution and use temperature as a conditional variable, with posterior-only tempering where we do not temper the base distribution, and standard variational inference with a normalizing flow model (NF VI). We also include linear annealing~\cite{rezende2015variational} and AdaAnn~\cite{cobian2023adaann}, where the temperature schedule of annealing is adaptively determined based on local estimates of KL divergence between intermediate target distributions. All methods are given the same number of training epochs; as such, the AdaAnn hyperparameters are tuned such that the annealing schedule reaches $T=1$ before the training ends. 

We find that our method is relatively insensitive to the tempering hyperparameters, and thus sample temperatures uniformly in the range $T\in[0.95, 10]$ when training. Models are trained for $10^4$ epochs using either this temperature range, when using a conditional tempering method, or the respective annealing schedule. Following that, models are fine tuned for $5\times10^3$ epochs, sampling uniformly in the interval $T\in[0.95, 1.5]$ or using $T=1$ for the conditional and non-conditional methods, separately. For both linear annealing and AdaAnn, we adopt a temperature range of $T \in [1, 100]$, consistent with prior work~\cite{rezende2015variational, cobian2023adaann}. All normalizing flow models are implemented using \texttt{flowjax}~\cite{daniel_ward_2024_11204962} as 10-layer Rational-Quadratic neural spline flows with coupling layers~\cite{Durkan2019NeuralSF}, with 16 knots and an interval width of 8. Each coupling layer is parameterized by a dense neural network with 5 hidden layers of width 1024 and \texttt{silu} activations~\cite{hendrycks2023gaussianerrorlinearunits, Elfwing2017SigmoidWeightedLU, ramachandran2017searchingactivationfunctions}. The base distribution is a standard multivariate Gaussian over the latent space. Additional hyperparameters can be found in~\cref{sec:hyperparameters}.



\begin{figure}[htb]
  \centering
  \includegraphics[width=\textwidth]{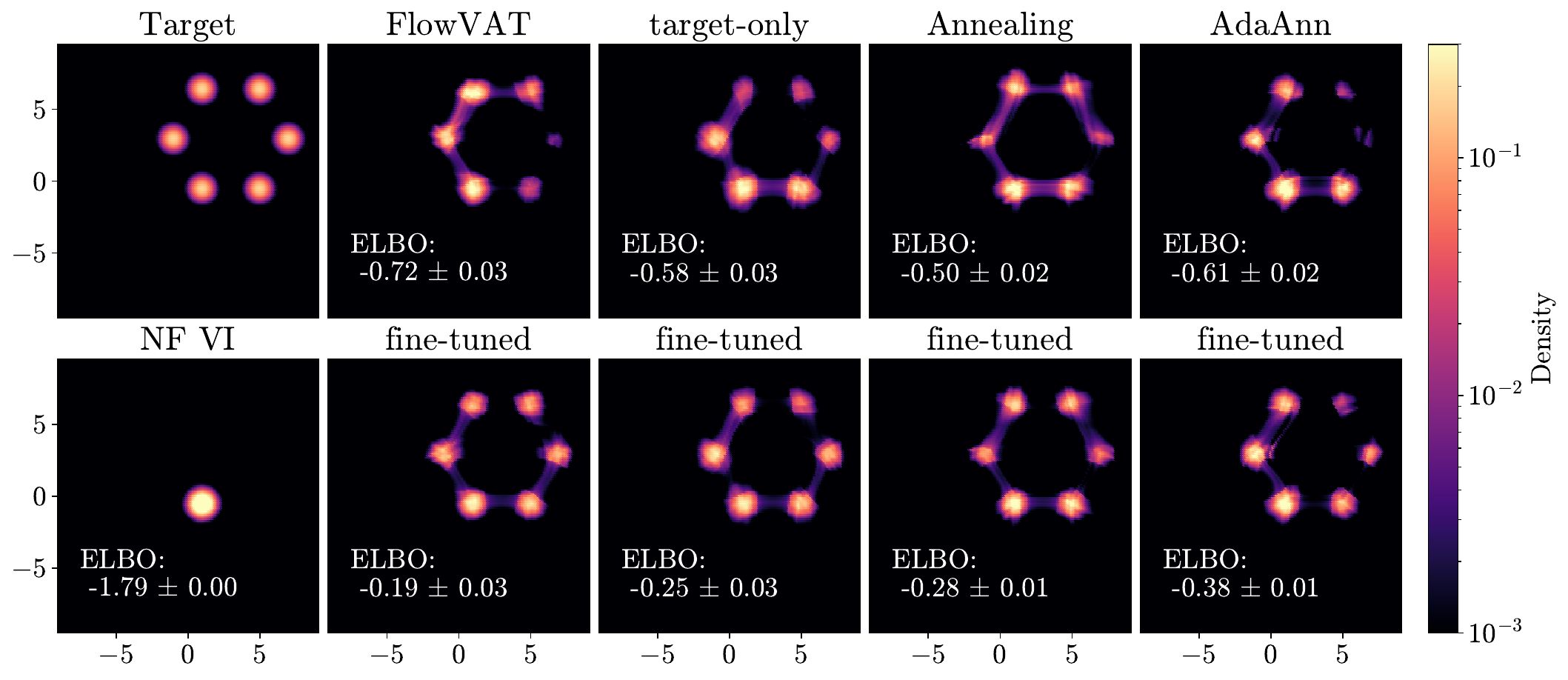}
  \caption{Comparison of different variational inference methods on 2D multimodal targets. The top row shows the target distribution and results from FlowVAT (this work), target-only tempering, traditional annealing, and AdaAnn. The bottom row includes normalizing flow VI and fine-tuned versions of each method, where additional training epochs are run at either $T=1$ for the annealed methods or $T\in[0.95, 1.5]$ for the conditional tempering methods including FlowVAT. ELBO values with standard deviations are shown for each approach. We can see that our approach, FlowVAT, performs better than all other methods.}
  \label{fig:2D_gaussians}
\end{figure}

\subsection{2D multimodal targets}\label{ssec:2D_exp}

\begin{figure}[ht]
  \centering
  \includegraphics[width=0.5\textwidth]{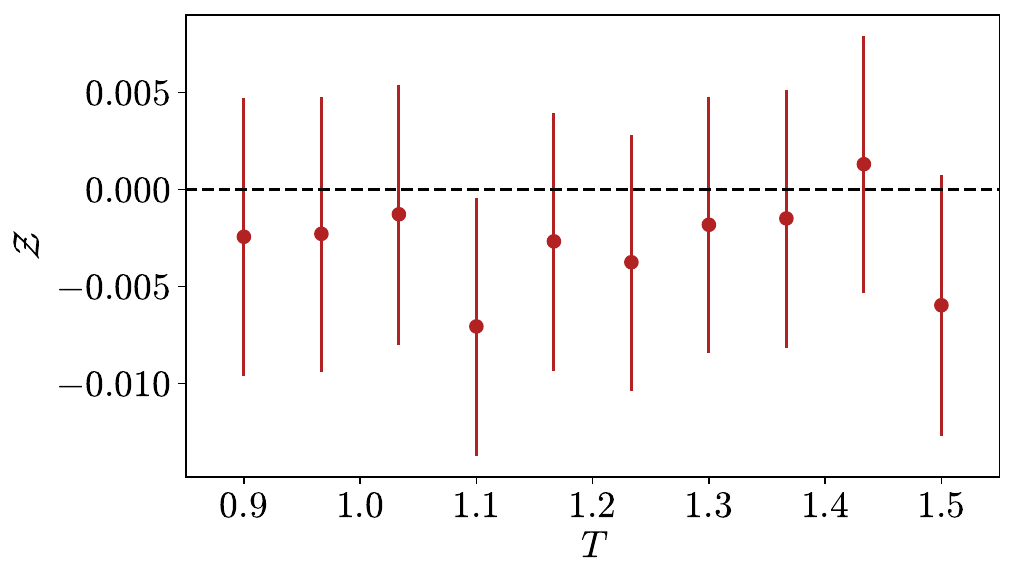}
  \caption{Estimated evidence via importance sampling (see~\cref{eq:IS_evidence}) for the 2D multimodal target posterior along with $1\sigma$ errorbars. As the true posterior is well approximated by the variational posterior, we do not expect a strong temperature dependence.}
  \label{fig:2D_Z}
\end{figure}

In this experiment, we evaluate the ability of each algorithm to find separated modes in a 2D space. The target posterior has 6 Gaussian modes distributed in a ring; details of this distribution are given in~\cref{sec:2D_target_parameters}. This target posterior, along with the variational posteriors after initial training and fine-tuning, are shown in~\cref{fig:2D_gaussians}. 

We find that all methods with tempering or annealing perform well after fine-tuning and finds all modes, while normalizing flow VI without tempering fails to find isolated modes. In addition, our conditional tempering method, FlowVAT, does not need the tempering hyperparameter to be tuned, presenting a key advantage for black-box VI, and also provides a slightly tighter ELBO. 

We additionally estimate the true evidence using $5\times10^4$ importance samples, as described in~\cref{eq:IS_evidence}, with the FlowVAT variational posterior at various temperatures in the range $T\in[0.95, 1.5]$, as shown in~\cref{fig:2D_Z}. We find that the estimated values are consistent with the ground truth, $\mathcal{Z} = 0$. We do not observe a strong temperature dependence, as the variational posterior approximates the true posterior well, and thus even at temperatures slightly below one there is good coverage of the posterior.

\begin{figure}[htb]
  \centering
  \includegraphics[width=\textwidth]{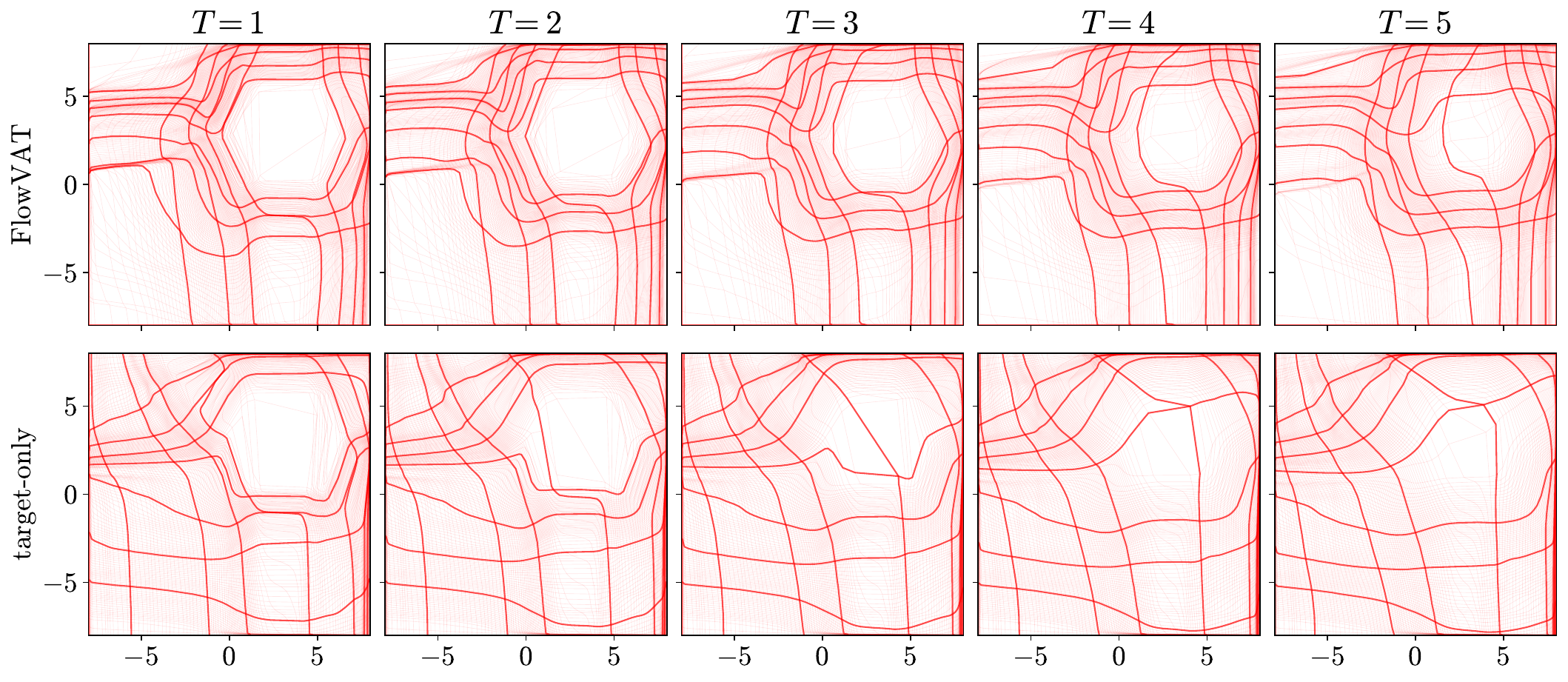}
  \caption{The transformation represented by the normalizing flow at various temperatures for the FlowVAT model (top), or target-only tempering (bottom). These plots are produced by transforming a regular grid using the normalizing flow models; as such, an identity transformation would be represented by a regular grid with a grid-spacing of 2. Models without fine-tuning are used to produce this diagram as we are including temperatures outside of the fine-tuning temperature range.}
  \label{fig:2D_transforms}
\end{figure}

Finally, we can examine whether our theoretical arguments from~\cref{sec:flowvat} regarding affine-invariant tempering are reflected in the trained flows. In particular, we would expect that the transformation changes much less as temperature varies when both the base and the target distributions are tempered, as is done with FlowVAT, as shown in~\cref{fig:2D_transforms}. We can see that the FlowVAT model indeed produces a normalizing flow where the transformations change much less with temperature when compared with target-only tempering, where the model has to fully adapt to the tempered posterior distribution.

\subsection{Higher dimensional multimodal targets}
\label{sec:highd}

To evaluate mode discovery performance in challenging settings, we construct synthetic Gaussian mixture posteriors (GMs) in 10 and 20 dimensions. Each mixture contains five equally weighted isotropic components with unit covariance. The mode locations are generated such that the pairwise distance between any two modes is at least the 0.99 quantile of the corresponding Gaussian, while ensuring that each mode lies within the 0.999 quantile of at least one other mode. This setup is designed to mimic the clustered multimodal structures often observed in scientific posteriors, such as those in neutrino oscillation analyses~\cite{esteban2018updated, coloma2023global}, searches for physics beyond the Standard Model~\cite{Fowlie:2011mb}, and gravitational wave inference~\cite{Dax:2024mcn}.

The entire mixture is normalized so that the true log-evidence is zero. For each dimension, we generate 10 randomized instances of the GM and evaluate performance using 2000 samples from the variational posterior per trial. A mode is considered ``captured'' if more than 5\% of the samples fall within the 0.9 quantile region of the corresponding Gaussian component. In addition, we report the estimated ELBO.


\begin{table}[ht]
\centering
\caption{Comparison of methods across datasets using number of modes captured and ELBO. Higher is better for both. The target posteriors contain 5 modes each. For the 10D posterior, standard variational inference performs best in terms of the ELBO, but does not find all modes reliably. FlowVAT outperforms all other methods for the more challenging 20D posterior, while performing the best out of all tempered/annealed methods for the 10D posterior.}
\label{tab:modes_elbo}
\begin{tabular}{lcccc}
\toprule
 & \multicolumn{2}{c}{\textbf{GM-10D}} & \multicolumn{2}{c}{\textbf{GM-20D}} \\
\cmidrule(lr){2-3} \cmidrule(lr){4-5}
\textbf{Method} & \textbf{Modes} & \textbf{ELBO} & \textbf{Modes} & \textbf{ELBO} \\
\midrule
NF VI     & $4.5\pm0.5$ & $\mathbf{-0.25\pm0.07}$ & $1.0\pm0.0$ & $-1.6\pm0.0$ \\
\textbf{FlowVAT (this work)}  & $\mathbf{5.0\pm0.0}$ & $-0.45\pm0.05$  & $\mathbf{3.4\pm0.5}$ & $\mathbf{-1.2\pm0.2}$ \\
Target-only     & $\mathbf{5.0\pm0.0}$ & $-0.55\pm0.06$  & $3.0\pm0.7$ & $-1.5\pm0.2$ \\
Annealing       & $4.7\pm0.4$ & $-0.68\pm0.08$  & $1.8\pm0.6$ & $-1.6\pm0.2$ \\
AdaAnn          & $\mathbf{5.0\pm0.0}$ & $-0.49\pm0.08$ & $2.1\pm0.7$ & $-1.7\pm0.2$ \\
\bottomrule
\end{tabular}
\end{table}

Results are summarized in~\cref{tab:modes_elbo}. While no single method achieves uniformly superior performance across all metrics and dimensions, FlowVAT demonstrates strong and consistent performance. In 20 dimensions, it captures the most modes on average and achieves the highest ELBO, indicating a better fit to the target distribution. In 10 dimensions, where mode discovery is easier--as indicated by NF VI recovering most modes and achieving the highest ELBO--several methods including FlowVAT, target-only tempering, and AdaAnn successfully recover all modes.

\begin{figure}[ht]
  \centering
  \includegraphics[width=0.49\textwidth]{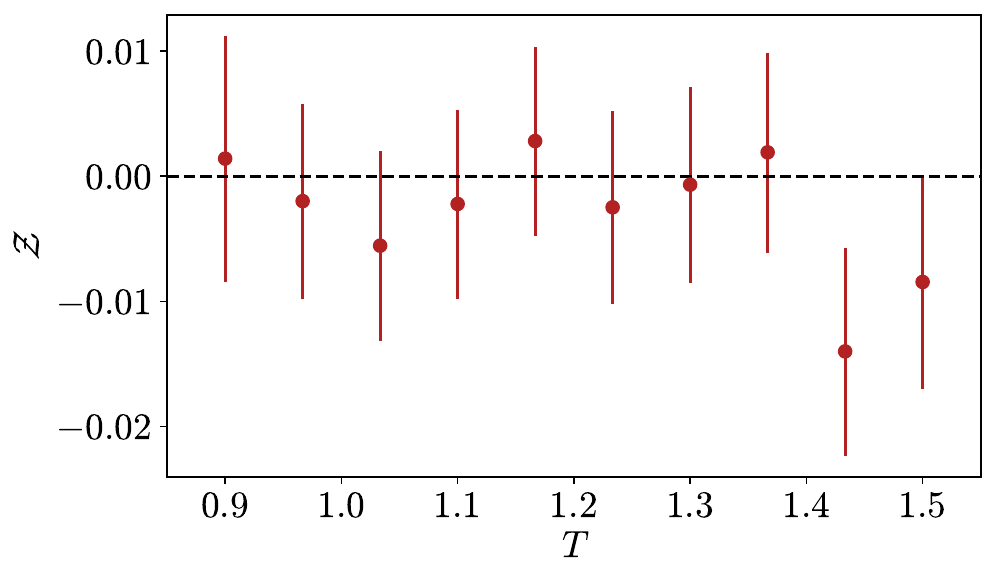} \includegraphics[width=0.49\textwidth]{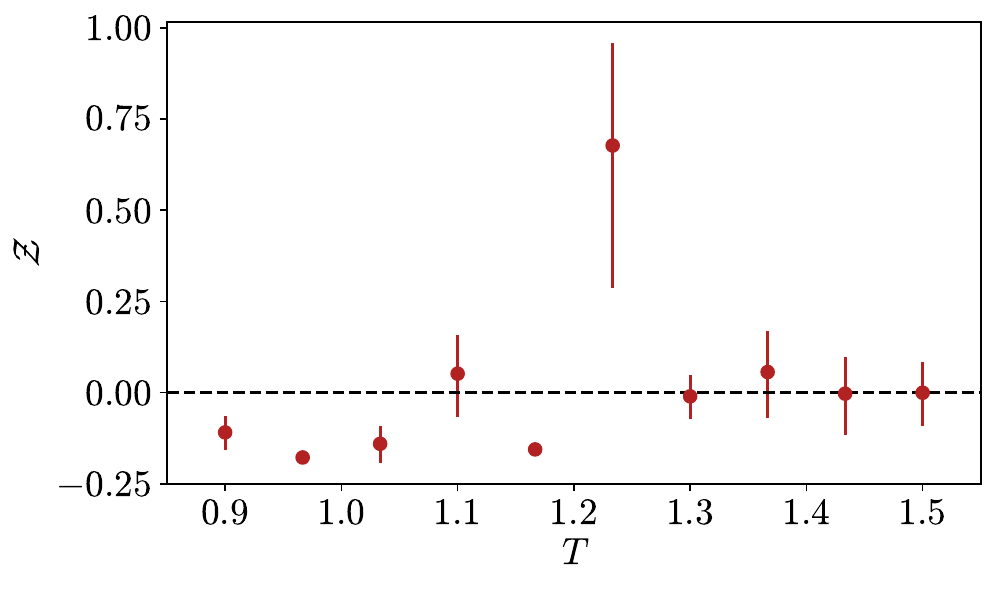}
  \caption{Estimated evidence via importance sampling (see~\cref{eq:IS_evidence}) for instances of the 10D (left) and 20D (right) multimodal target posteriors along with $1\sigma$ errorbars. As the true posterior is well approximated by the variational posterior in the 10D case, there is no temperature dependence. In the more challenging 20D case, we can see that the estimated evidence converges for $T\geq1.3$.}
  \label{fig:nD_Z}
\end{figure}

We hypothesize that FlowVAT’s advantage in higher dimensions stems from its affine-invariant tempering design, which reduces the burden on the flow model to learn complicated geometric transformations. In $d$ dimensions, a full affine transformation involves $d(d+1)$ parameters. By tempering both the target and base distributions in an affine-invariant manner, FlowVAT allows the normalizing flow to focus on modeling the residual non-affine structure, rather than spending capacity on learning global affine transformations. This leads to improved performance, particularly in higher dimensions where such transformations become more complex. Additional implementation details for this experiment are provided in Appendix~\ref{appendix:highd}.

We then estimate the Bayesian evidence using FlowVAT for one randomly chosen example for the 10 and 20 dimensional posteriors each using $5\times10^4$ importance samples. This can be seen in~\cref{fig:nD_Z}. We can see that in both cases, we can obtain values that are consistent with the ground truth of zero. In the 10 dimensional case, the computed evidence does not have a significant dependence on temperature, similar to the two dimensional case in~\cref{ssec:2D_exp}. In 20 dimensions, however, as the variational posterior is a poorer approximation of the target posterior, the computed evidence shows a clear bias at low temperature, with data points more than two standard deviations away from the truth values. Despite this, the evidence computed via importance sampling is stabilized for $T\geq1.3$, once again giving values compatible with the ground truth of zero.

\subsection{Benchmarks using eight-schools model}






\begin{table}[htb]
\centering
\caption{Evidence Lower Bound (ELBO) comparison on the Eight Schools model. Higher is better.}
\label{tab:elbo_bayes}
\begin{tabular}{lc}
\toprule
\textbf{Method} & \textbf{ELBO}  \\
\midrule
NF VI      & $-31.337\pm0.002$ \\
\textbf{FlowVAT (this work)}   & $-31.365\pm0.048$ \\
Target-only      & $-31.483\pm0.050$ \\
Annealing        & $-31.371\pm0.005$ \\
AdaAnn           & $-31.344\pm0.003$ \\
\bottomrule
\end{tabular}
\end{table}

We then evaluate all methods on the non-centered Eight Schools model. The model is implemented in \texttt{NumPyro}~\cite{phan2019composable, bingham2019pyro} following the formulation in \texttt{posteriordb}~\cite{magnusson2024posteriordb}, serving as a benchmark for assessing posterior approximation quality.

\begin{figure}[ht]
  \centering
  \includegraphics[width=\textwidth]{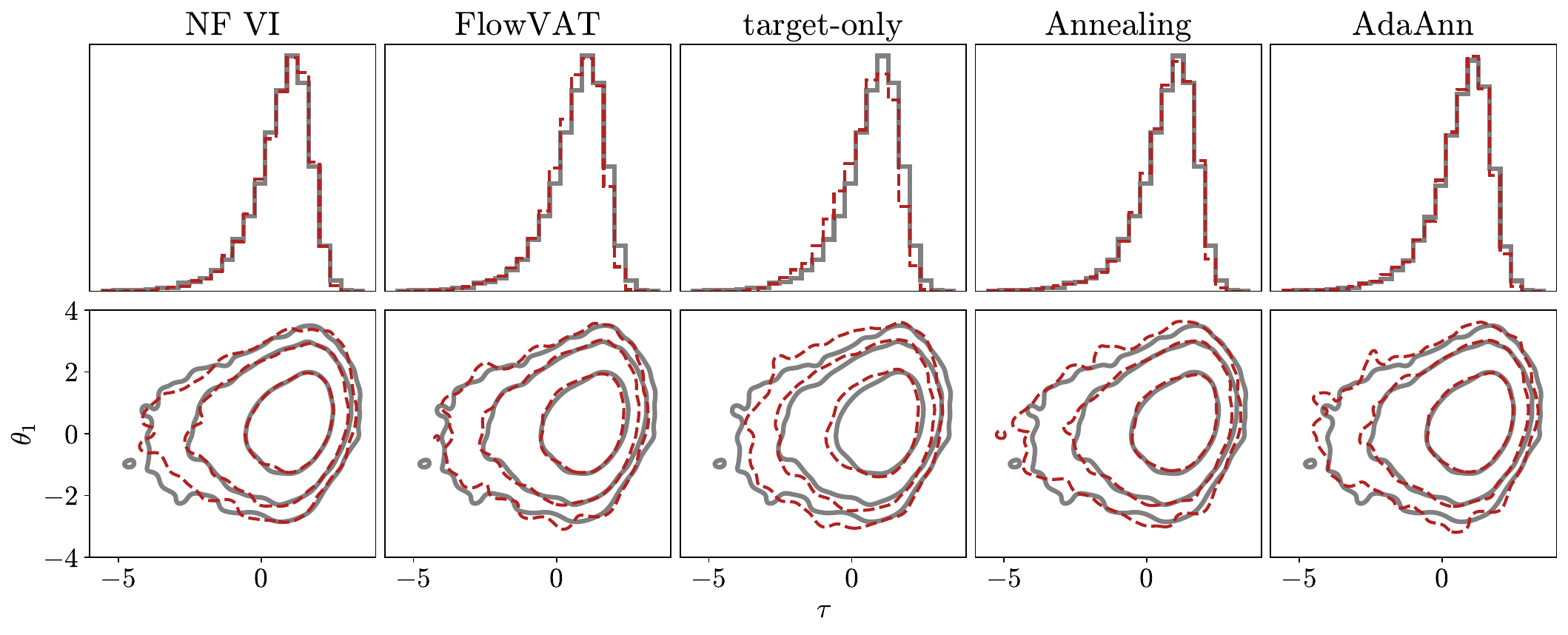}
  \caption{Comparison of variational inference methods on the eight schools model. Top row: Marginal histograms of $\theta_1$. Bottom row: Contour plots of the joint distribution between $\theta_1$ and $\tau$; additional plots are shown in~\cref{sec:eight_schools_plots}. Contours represent the 68\%, 95\%, and 99\% credible regions, with thick gray lines for the reference posterior and dashed red lines for each variational approximation.}
  \label{fig:eight_schools}
\end{figure}

We run a single training run per method and estimate the ELBO using 5000 samples. The reported uncertainties in~\cref{tab:elbo_bayes} correspond to the standard error of the Monte Carlo estimate. To ensure adequate support coverage for this model, we set the spline transformation interval to 32 during training. All methods achieve similar ELBO values, and visual inspection of the posterior approximations (\cref{fig:eight_schools}) shows close agreement with the reference across both marginal and joint distributions.

Among the methods compared, tempering only the target distribution performs slightly worse in terms of both ELBO and posterior shape. This shows that ablating the tempering of the base distribution adversely affects performance, and is consistent with our hypothesis that jointly tempering the base and target in an affine-invariant way reduces the burden on the flow to learn affine transformations.

\section{Discussion}

Our experiments demonstrate that FlowVAT offers substantial advantages over existing variational inference methods, particularly with challenging posteriors. We discuss the key implications of these results and their broader impact.

The ability to handle complicated multi-modal posteriors in high dimensions represents a significant advancement in variational inference~\cite{dhaka2021challenges}. Traditional variational methods frequently struggle with this combination, often exhibiting mode-seeking behaviour even when using expressive variational families like normalizing flows~\cite{dhaka2021challenges}. By simultaneously tempering both the base distribution and the target posterior, FlowVAT effectively addresses this fundamental limitation without the need for an annealing schedule. As demonstrated in~\cref{sec:experiments}, our approach captures more modes and achieves better ELBO values than competing methods, with this advantage becoming more pronounced as dimensionality increases. Such high-dimensional and multimodal posteriors are often encountered in science, including in neutrino oscillation analyses~\cite{esteban2018updated, coloma2023global}, searches for physics beyond the Standard Model~\cite{Fowlie:2011mb}, and gravitational wave inference~\cite{Dax:2024mcn}.

Beyond the performance improvements, FlowVAT represents a meaningful step toward truly hyperparameter-free black-box variational inference. A persistent challenge in practical applications of variational inference has been the need to carefully tune hyperparameters~\cite{agrawal2020advances}, particularly for difficult posteriors that require annealing. The requirement for such manual tuning undermines the "black-box" nature that makes variational inference attractive as a general-purpose inference method. Instead of requiring problem-specific schedule tuning, temperature is given to FlowVAT as a conditional variable, allowing the model to learn appropriate transformations across the entire temperature spectrum.

Our approach also introduces a beneficial inductive bias through affine-invariant tempering. As we argue for in~\cref{sec:flowvat}, this approach promotes consistent matching of the first two moments across temperatures. This property is particularly valuable in real-world inference problems where the target distribution may have complex structure and be located far from the origin in parameter space. The combination of this inductive bias with the generalization capabilities of overparameterized neural networks enables our model to efficiently transfer knowledge between different temperature regimes. In particular, we observe that ablation of the base distribution tempering adversely affects performance across all test cases, in support of our theoretical arguments.

In addition, the ability to sample from the variational posterior with a conditional temperature enables more reliable evidence estimation through importance sampling. Estimation of Bayesian evidence is useful for model comparison and hypothesis testing, but is traditionally very computationally intensive~\cite{Srinivasan:2024uax}. As demonstrated in our evidence computation experiments in~\cref{sec:experiments}, FlowVAT enables fast and stable importance sampling of the evidence integral, and thus has the potential to accelerate model comparison and hypothesis testing as well.

In summary, FlowVAT addresses three key challenges in variational inference: capturing complicated multi-modal structure in high dimensions, enabling stable estimation of the Bayesian evidence, and reducing dependency on annealing hyperparameters. These advances bring us closer to the goal of developing truly automatic black-box variational inference methods capable of handling the diverse and challenging posteriors encountered in real-world applications.

\subsection{Limitations and future work}
One key limitation is that we have only explored inference in a continuous parameter space, as normalizing flows reparameterize such continuous spaces. Future work could explore architectural innovations needed to use such an approach when some parameters are discrete, for example by combining a conditional normalizing flow with a learned categorical distribution. In addition, we did not conduct experiments with large imbalances in the probability mass assigned to various modes, as capturing multiple modes is most crucial when modes have comparable probability mass. Finally, in this work, we focused on introducing the key ideas; while we are releasing the code used to produce experiments in this paper, we hope to release a user friendly module in the future.






\section{Acknowledgment}

We thank Ivy Li for his comments on the manuscript. This work is supported by the Department of Energy AI4HEP program and the National Science Foundation CAREER award PHY-204659.

{
\small
\bibliographystyle{unsrt}
\bibliography{biblio}
}


\appendix

\section{Broader Impacts}

FlowVAT represents an advance in variational inference, which has seen significant use in scientific research. As such, we expect that improving the performance of variational inference will create positive societal impacts by accelerating research and scientific discovery. We do not expect this work to produce negative societal impacts beyond knock-on effects from research that uses our methods.

\section{Experiments and Details}

\subsection{Computational resource}
\label{appendix:resource}
All experiments were run in a containerized environment with 8 CPU cores and 64~GB of RAM. A single NVIDIA A40 GPU (48~GB memory) was used for all experiments.  The training time remains similar ($\sim18$ min) for all the methods, which is expected as the model architecture stays unchanged.

\subsection{Experiment hyperparameters}\label{sec:hyperparameters}
\begin{table}[ht]
\centering
\caption{Additional hyperparameters that were not included in the main text.}
\label{tab:hyperparams}
\begin{tabular}{lcccc}
\toprule

\textbf{Hyperparameter} & \textbf{2D}   & \textbf{10D} & \textbf{20D}& \textbf{Eight schools} \\
\midrule
Optimizer               & \texttt{adamW} & \texttt{adamW} & \texttt{adamW}     & \texttt{adamW}\\
Pre-training learning rate&$5\times10^{-6}$& $5\times10^{-6}$& $5\times10^{-6}$ & $5\times10^{-6}$\\
fine-tuning learning rate&$10^{-6}$      & $10^{-6}$ & $10^{-6}$         & $10^{-6}$\\
AdaAnn tolerance        & $0.02$ & $0.1$ & $0.15$ & $0.1$ \\
\bottomrule
\end{tabular}
\end{table}

For the linear annealing and AdaAnn methods, the temperature $T$ is adjusted every 100 epochs.

\subsection{2D multimodal distribution parameters}\label{sec:2D_target_parameters}
The 2D multimodal distribution is centered around $[3,3]$, with 6 evenly spaced modes placed on a ring. The radius of the ring is $4$, and the standard deviation of each mode is $0.38$, chosen such that NF VI consistently fails to find all but one mode.

\subsection{Target posterior generation for high dimension multimodal experiment}
\label{appendix:highd}

The generation of the high dimensional multimodal targets rely on the quantiles of multivariate normal distribution. In \( d \)-dimensional space, the squared Euclidean norm of a sample from \( \mathcal{N}(0, \sigma^2 I_d) \) follows a scaled chi-squared distribution:
\[
\|x\|^2 \sim \sigma^2 \cdot \chi^2_d.
\]
To define a threshold distance that contains a specified quantile (e.g., 95\%) of the Gaussian mass, we compute the square root of the corresponding chi-squared quantile:
\[
\delta_{\text{quantile}} = \sigma \cdot \sqrt{F^{-1}_{\chi^2_d}(\alpha)},
\]
where \( F^{-1}_{\chi^2_d}(\alpha) \) is the inverse CDF (quantile function) of the chi-squared distribution with \( d \) degrees of freedom, and \( \alpha \in (0,1) \) is the desired quantile level (e.g., \( \alpha = 0.95 \)).

This threshold is used to set reasonable values for \texttt{min\_dist} or \texttt{max\_dist} when sampling Gaussian centers in \( \mathbb{R}^d \), ensuring that the modes are neither too tightly clustered nor overly dispersed.
As mentioned in \cref{sec:highd}, a 0.99 quantile is used as \texttt{min\_dist} and a 0.999 quantile is used as \texttt{max\_dist}. The Gaussian centers are generated by sampling points in a hypercube using the algorithm described in Algorithm~\ref{alg:generate_centers}.

\begin{figure}[ht]
  \centering
  \includegraphics[width=\textwidth]{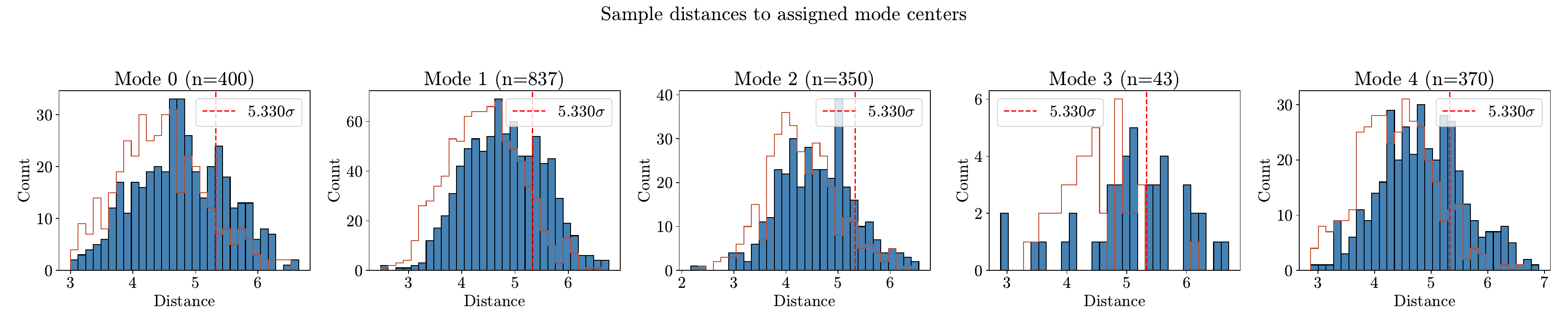}
  \caption{An example of visual check of the training result for FlowVAT on 20D Gaussian mixture distribution. The distribution of absolute distance between the center of the modes and the assigned samples is shown for each mode. The blue histogram represents the samples from the normalizing flow model and the red steps histogram represents the expected distribution from a 20D Gaussian distribution. The dashed red vertical line marks the 0.9 quantile. Here, Mode 3 is considered not captured as the ratio of the samples within the threshold is smaller than 0.05.}
  \label{fig:20_d_example}
\end{figure}

A preliminary visual check is done to inspect if the model fits the multimodal target. An example is shown in \cref{fig:20_d_example}. Each sample from the normalizing flow model is assigned to the closest center, and the distribution of the absolute distance to the center is compared with the one from a Gaussian distribution.

\begin{algorithm}[ht]
\caption{Generation of Gaussian Centers}
\label{alg:generate_centers}
\begin{algorithmic}[1]
\REQUIRE Number of centers $K$, dimension $d$, minimum distance $d_{\min}$, maximum distance $d_{\max}$, batch size $M$, random seed
\ENSURE Set of $K$ well-separated Gaussian centers $\mathcal{C} = \{c_1, \dots, c_K\} \subset \mathbb{R}^d$

\STATE Initialize random number generator with given seed
\STATE Sample initial center $c_1 \sim \mathcal{U}([-5, 5]^d)$
\STATE $\mathcal{C} \gets \{c_1\}$
\WHILE{$|\mathcal{C}| < K$ \AND max\_tries not exceeded}
    \STATE Sample $M$ candidates $\{x_1, \dots, x_M\} \sim \mathcal{U}([-5, 5]^d)$
    \FOR{each candidate $x_i$}
        \STATE Compute distances $D_i = \{\|x_i - c\| : c \in \mathcal{C}\}$
        \IF{$\min D_i > d_{\min}$ \AND $\min D_i < d_{\max}$}
            \STATE $\mathcal{C} \gets \mathcal{C} \cup \{x_i\}$
            \IF{$|\mathcal{C}| = K$}
                \STATE \textbf{break}
            \ENDIF
        \ENDIF
    \ENDFOR
\ENDWHILE
\RETURN $\mathcal{C}$
\end{algorithmic}
\end{algorithm}

\subsection{Posterior plots for the eight schools experiment}\label{sec:eight_schools_plots}

\begin{figure}[ht]
  \centering
  \includegraphics[width=0.7\textwidth]{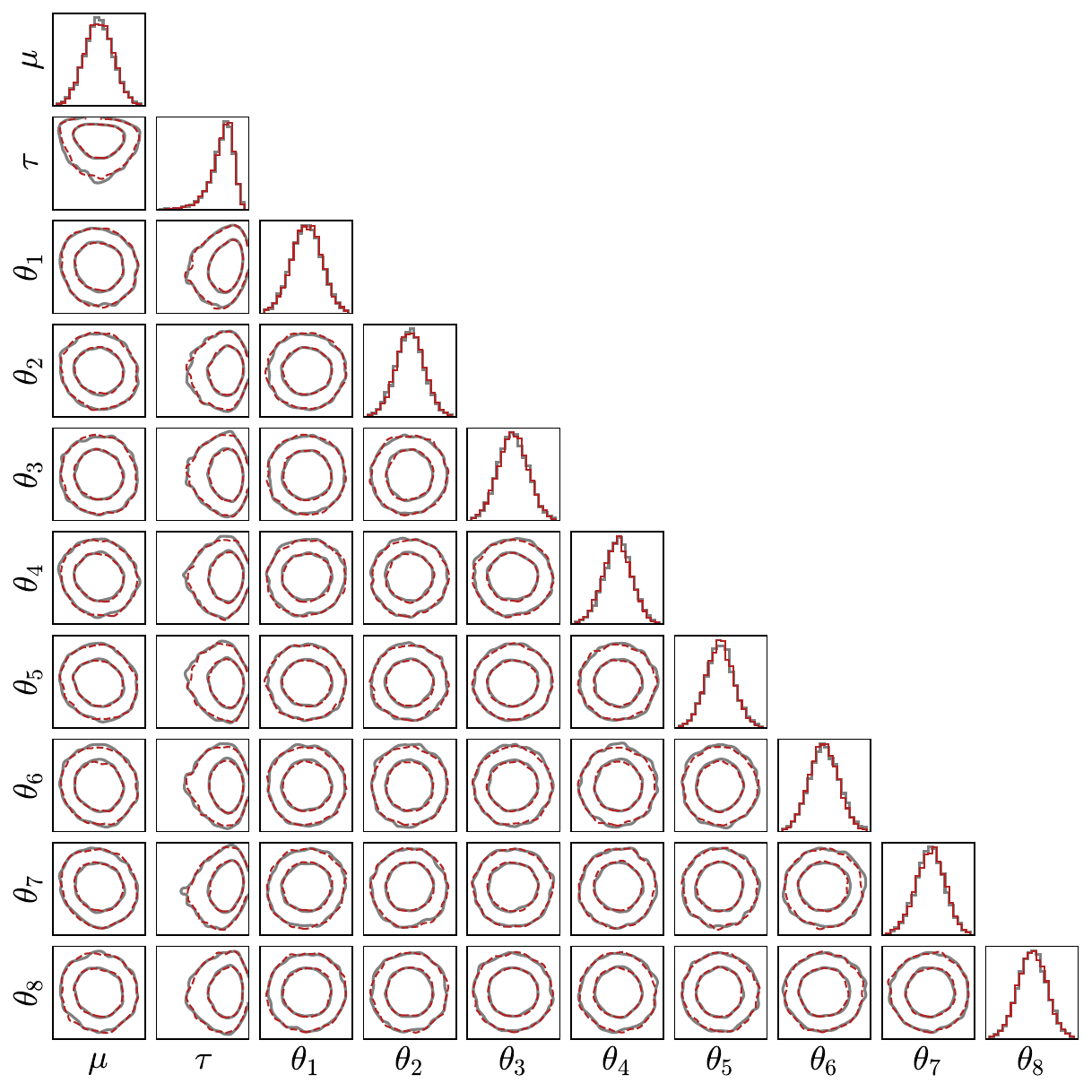}
  \caption{Full corner plot of NF VI for the Eight Schools model}
  \label{fig:eight_schools_full_flow}
\end{figure}

\begin{figure}[ht]
  \centering
  \includegraphics[width=0.7\textwidth]{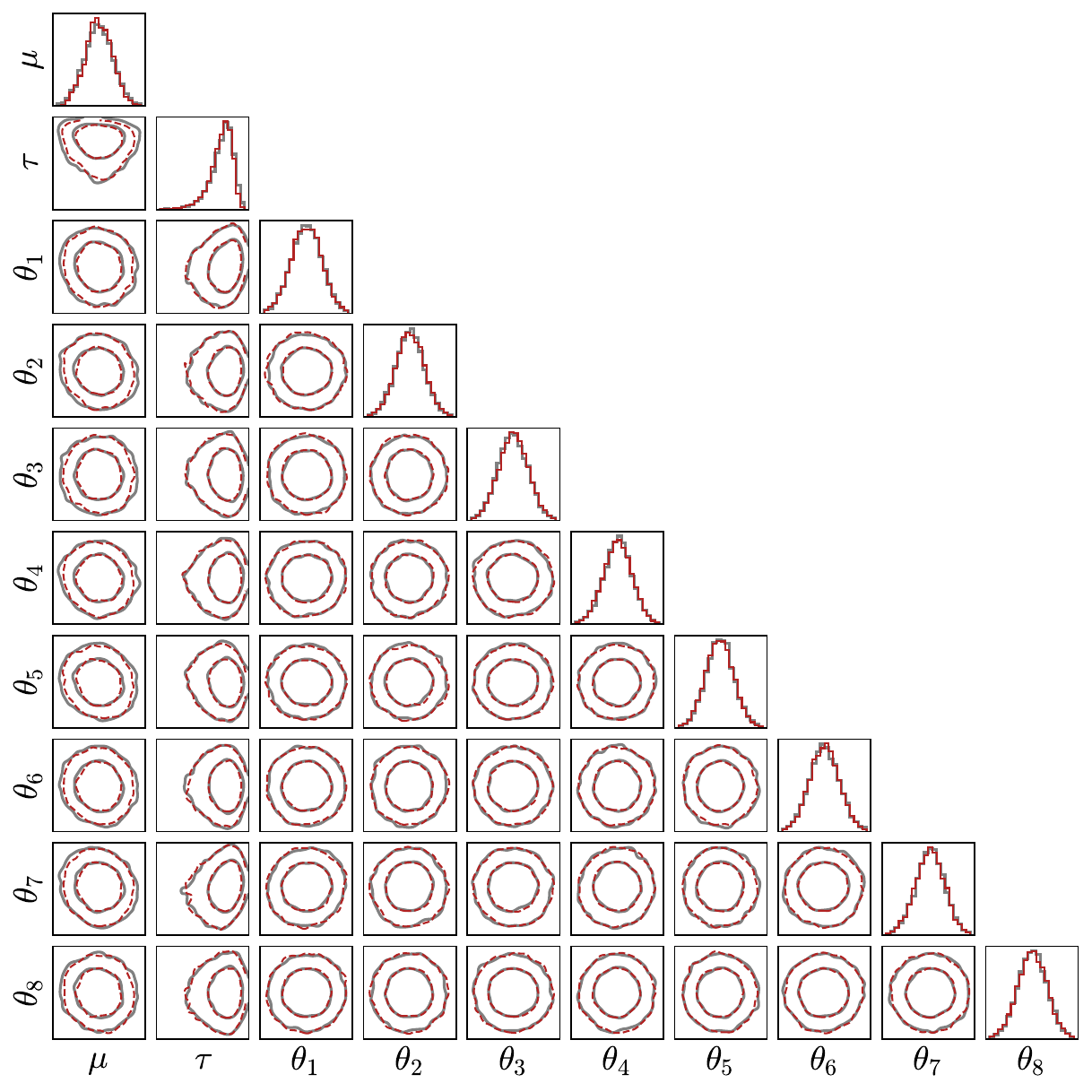}
  \caption{Full corner plot FlowVAT for Eight Schools model}
  \label{fig:eight_schools_full_flowvat}
\end{figure}

\begin{figure}[ht]
  \centering
  \includegraphics[width=0.7\textwidth]{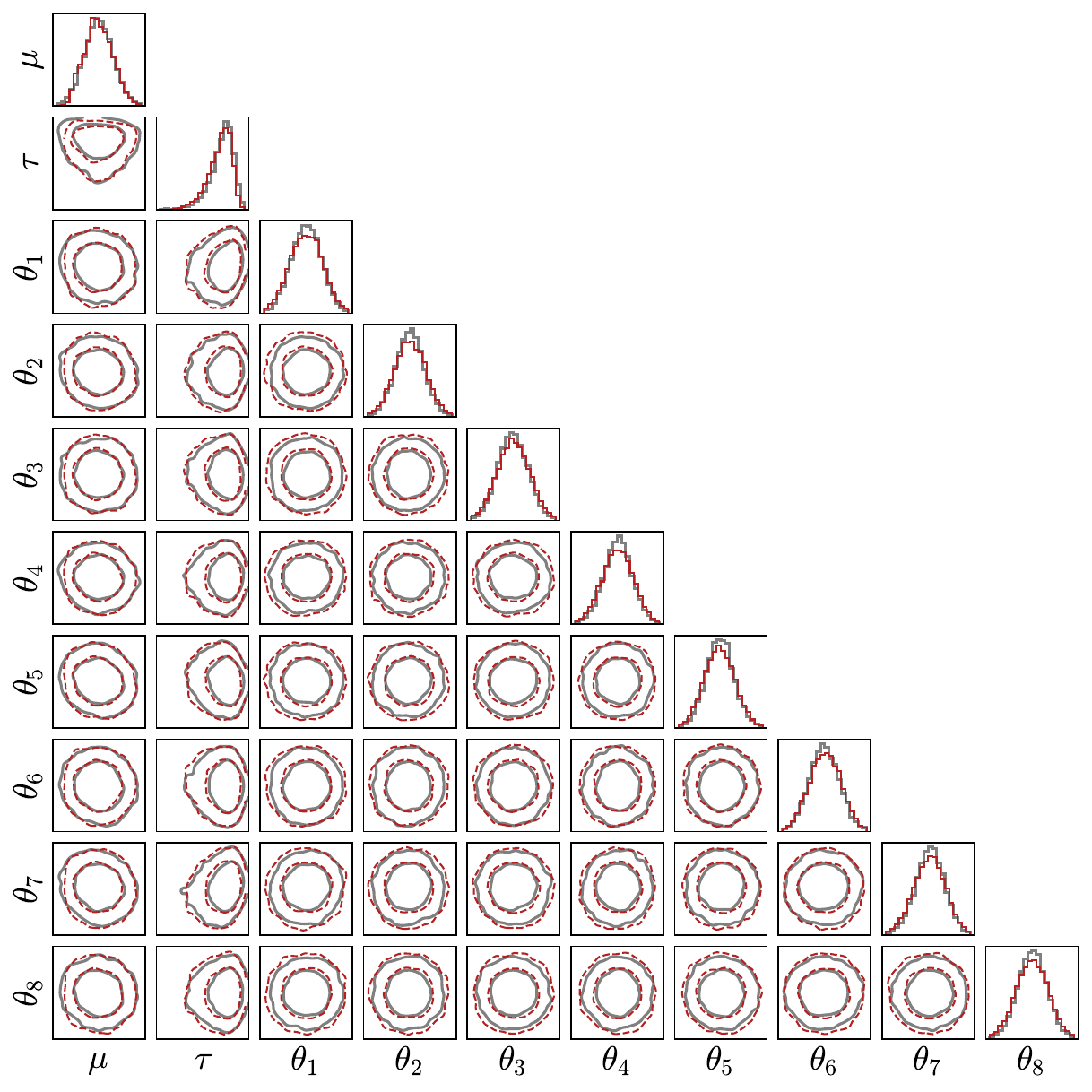}
  \caption{Full corner plot target-only tempering for Eight Schools model}
  \label{fig:eight_schools_full_target_only}
\end{figure}

\begin{figure}[ht]
  \centering
  \includegraphics[width=0.7\textwidth]{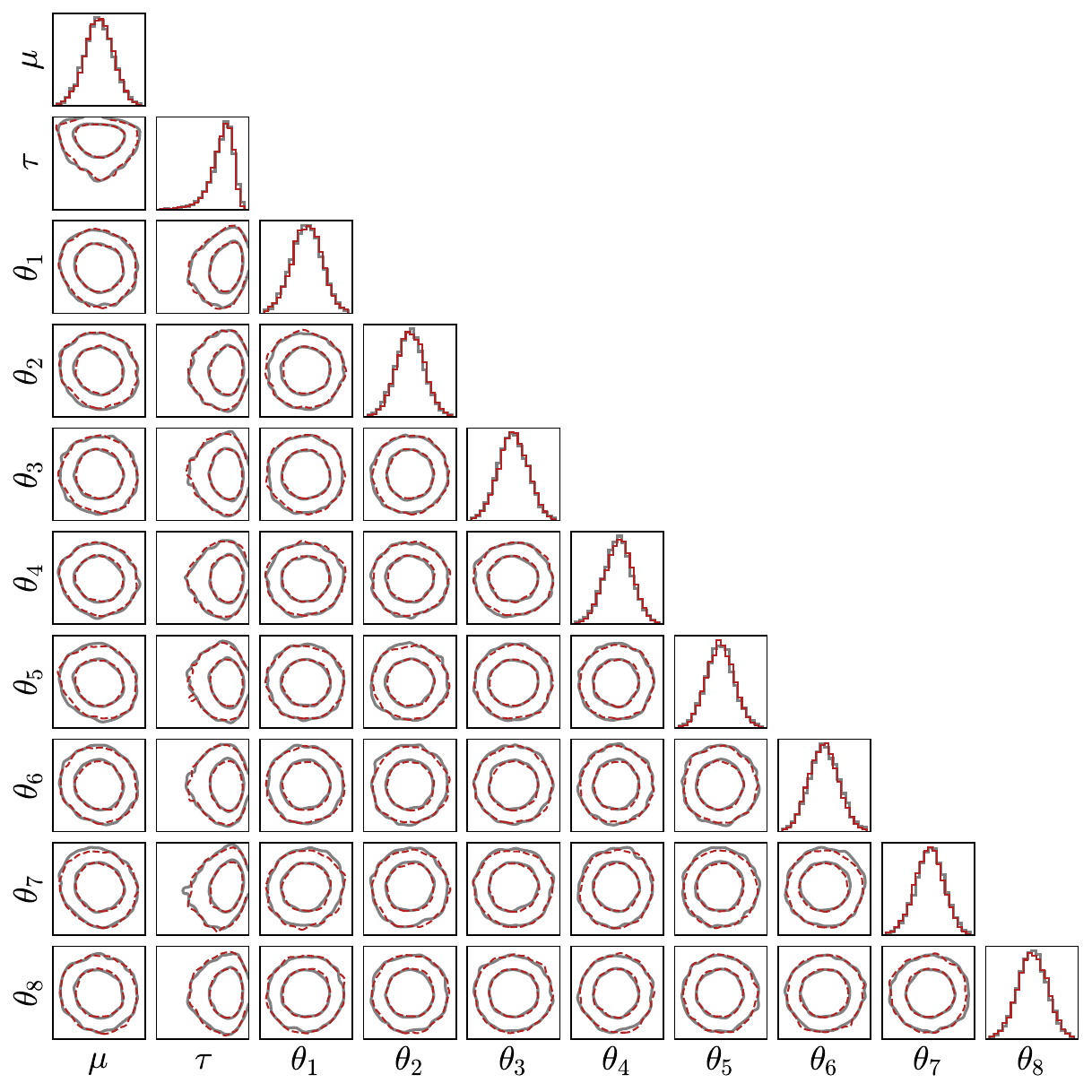}
  \caption{Full corner plot Annealing for Eight Schools model}
  \label{fig:eight_schools_full_linann}
\end{figure}

\begin{figure}[ht]
  \centering
  \includegraphics[width=0.7\textwidth]{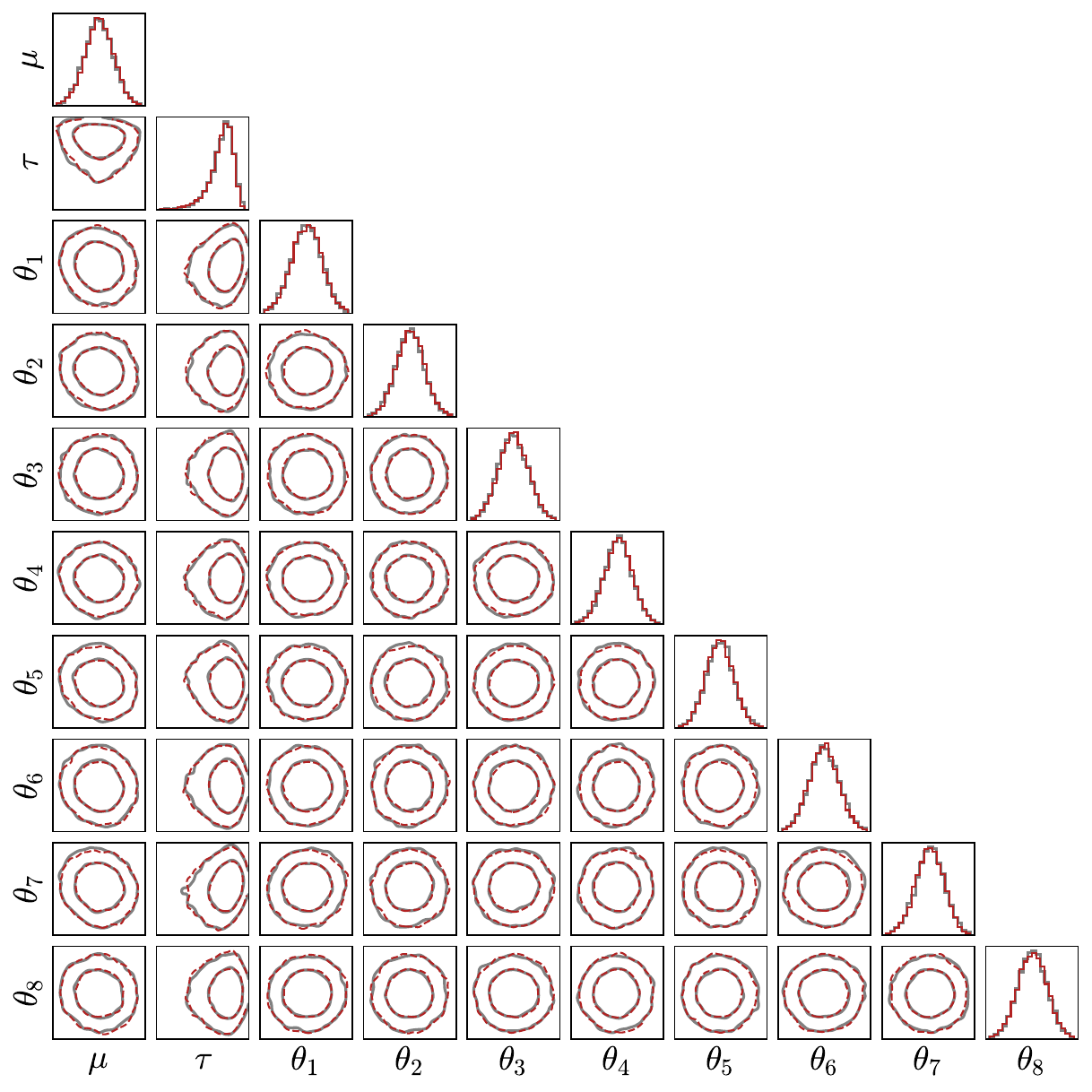}
  \caption{Full corner plot AdaAnn for Eight Schools model}
  \label{fig:eight_schools_full_adaann}
\end{figure}

In the full corner plots of the inferred posterior versus reference posterior (\cref{fig:eight_schools_full_flow}, \cref{fig:eight_schools_full_flowvat}, \cref{fig:eight_schools_full_target_only}, \cref{fig:eight_schools_full_linann} and \cref{fig:eight_schools_full_adaann} ), contours represent the 68\% and 95\% credible regions. Thick gray lines represent the reference posterior and dashed red lines represent variational approximation.

\end{document}